\pdfoutput=1

\documentclass[preprint,12pt]{elsarticle}

\usepackage{times}
\usepackage{soul}
\usepackage{url}
\usepackage[hidelinks]{hyperref}
\usepackage[T1]{fontenc}
\usepackage[utf8]{inputenc}
\usepackage[small]{caption}
\usepackage{subcaption}
\usepackage{graphicx}
\usepackage{amsmath}
\usepackage{amsthm}
\usepackage{multirow}
\usepackage{mathtools}
\usepackage{enumerate}
\usepackage{booktabs}
\usepackage{varwidth}
\usepackage{latexsym}
\usepackage{amssymb}
\usepackage{booktabs}
\usepackage{amsfonts}
\usepackage{amsmath}
\usepackage{color}
\usepackage{ulem}
\usepackage[colorinlistoftodos]{todonotes}
\usepackage{enumitem}
\usepackage{latexsym}

\usepackage{microtype}

\begin{document}

\begin{frontmatter}

%
%

\title{
       Image Captioning for Effective Use of Language Models in Knowledge-Based Visual Question Answering}


\author[label1]{Ander Salaberria}
\ead{ander.salaberria@ehu.eus}
\author[label1]{Gorka Azkune\corref{cor1}}
\ead{gorka.azcune@ehu.eus}
\author[label1]{Oier Lopez de Lacalle}
\ead{oier.lopezdelacalle@ehu.eus}
\author[label1]{Aitor Soroa}
\ead{a.soroa@ehu.eus}
\author[label1]{Eneko Agirre}
\ead{e.agirre@ehu.eus}

\address[label1]{
HiTZ Basque Center for Language Technologies - Ixa NLP Group, University of the Basque Country (UPV/EHU), M. Lardizabal 1, Donostia 20018, Basque Country, Spain}


\begin{abstract}
Integrating outside knowledge for reasoning in visio-linguistic tasks such as visual question answering (VQA) is an open problem. Given that pretrained language models have been shown to include world knowledge, we propose to use a unimodal (text-only) train and inference procedure based on automatic off-the-shelf captioning of images and pretrained language models. 
Our results on a visual question answering task which requires external knowledge (OK-VQA) show that our text-only model outperforms pretrained multimodal (image-text) models of comparable number of parameters. In contrast, our model is less effective in a standard VQA task (VQA 2.0) confirming that our text-only method is specially effective for tasks requiring external knowledge. In addition, we show that increasing the language model's size improves notably its performance, yielding results comparable to the state-of-the-art with our largest model, significantly outperforming current multimodal systems, even though augmented with external knowledge.
Our qualitative analysis on OK-VQA reveals that automatic captions often fail to capture relevant information in the images, which seems to be balanced by the better inference ability of the text-only language models. Our work opens up possibilities to further improve inference in visio-linguistic tasks.

\end{abstract}

\begin{keyword}
Visual Question Answering \sep Image Captioning \sep  Language Models \sep Deep learning 



\end{keyword}

\end{frontmatter}


\section{Introduction}
Most visio-linguistic tasks are framed in such a way that all the necessary information to solve them is in the images and texts provided in the dataset. That is the case of visual question-answering (VQA) \cite{antol2015vqa} or visual entailment \cite{xie2019visual}. In addition, some tasks require access to external knowledge in order to solve them. In this work we dive in \emph{Outside Knowledge VQA} (OK-VQA) \cite{marino2019ok}, where the image content is not sufficient to answer the questions. Contrary to self-contained VQA tasks, which can be solved grounding images and text alone, these tasks require methods that leverage external knowledge resources and are able to do inference on that knowledge.

\begin{figure}[t]
\centering
\includegraphics[width=0.68\linewidth]{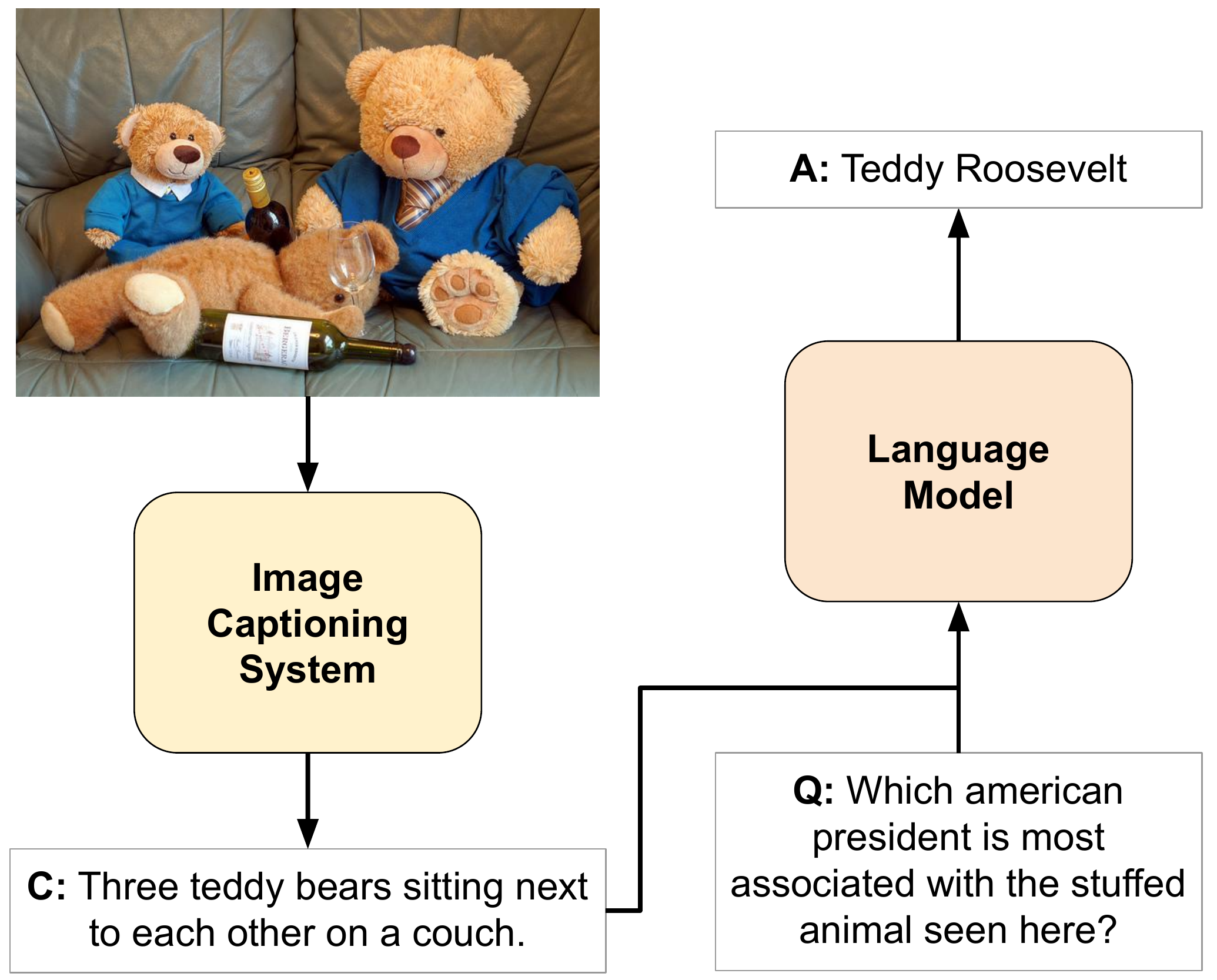}
\caption{Given a question and image, we verbalize the contents of the image and apply a pretrained language model for inference. We show that current text-only models are better in generalization and inference than multimodal models for knowledge-based VQA. 
} \label{system}
\end{figure}

External knowledge useful for OK-VQA can be broadly classified into two categories, according to \cite{marino2020krisp}: (i) symbolic knowledge, which can be represented using graphs, for example ConceptNet \cite{speer2017conceptnet}, and (ii) implicit knowledge, which is encoded in the weights of neural networks trained in different datasets. Supporting the later case, transformer-based language models (LM) pretrained in large corpora like BERT \cite{devlin2019bert} have been successfully used as implicit knowledge bases \cite{petroni2019language}. 

In this paper we focus on the use of implicit knowledge in the form of pretrained LMs. While using LMs is relatively common in OK-VQA, they are usually integrated into multimodal transformers by diverse means, so as to integrate the visual and textual inputs of the task. Given that LMs were originally designed to process textual input and are extensively trained in textual corpora, we hypothesized that a system that relies exclusively on text will allow LMs to better leverage their implicit knowledge. Because OK-VQA is a visio-linguistic task, we propose to use automatic image captioning as a way to verbalize the information in the image, where the captions are descriptions of the images which are used as input to the LMs. Once the captions are generated, all the inference in our method is done using text-only models. We are aware that captions do not contain all the information in an image, and want to check whether the text-only models can compensate for that initial loss of information. The approach proposed in this paper, named Caption-based Model or CBM can be seen in Figure~\ref{system}.

To validate our hypothesis, we present an extensive experimentation on the OK-VQA dataset, comparing our proposed caption-based model with the \textit{de facto} standard of visio-linguistic tasks, i.e. multimodal transformers, which are widely used in VQA tasks to process the questions (text) and images. We also focus on language models of different sizes, to see the impact of model capacity on OK-VQA. 
As a result of our experiments, we find that:
\begin{itemize} 
 \item Captions are more effective than images for OK-VQA when models of similar size are used as is, and achieve similar results when both are fine-tuned on additional VQA datasets. 
 \item Increasing the size and the capacity of language models allows to reach state-of-the-art results, outperforming by a large margin current multimodal transformers. Furthermore, we observe a trend of improvement that has not yet stabilized. 
 \item The complex use of in-context-learning as in PICa \cite{yang2021pica} does not beat fine-tuning our smaller model, that is, our system based on T5 \cite{raffel2020t5} obtains results comparable to an ensemble of five GPT-3 runs which are 15-times larger in parameters.

 \item The larger contribution of captions on OK-VQA with respect to results on a regular VQA dataset \cite{goyal2017vqav2} show that text-only systems are specially effective when external knowledge is needed. 
\end{itemize}

Our code is available at \textcolor{blue}{\href{https://github.com/salanueva/CBM}{https://github.com/salanueva/CBM}}.


\section{Related Work}

There are many \textbf{visual question-answering datasets} in the literature \cite{antol2015vqa, goyal2017vqav2, johnson2017clevr}, where given an image and a question about the contents of that image, a system has to provide a textual answer. Some VQA datasets also demand leveraging external knowledge to infer the answer and, thus, they are known as knowledge-based VQA tasks. Good examples are KB-VQA \cite{wang2017explicit}, KVQA \cite{shahMYP19}, FVQA \cite{wang2017fvqa} and OK-VQA \cite{marino2019ok}. KVQA requires knowledge about named entities (e.g. Barack Obama, White House, United Nations) and that knowledge is already provided as a graph. FVQA annotates questions by selecting a fact from a fixed knowledge base but its size is relatively small. KB-VQA is even smaller, presenting template-based questions whose answers can be obtained reasoning over commonsense resources or Wikipedia. In contrast, OK-VQA requires knowledge from unspecified external resources and, although smaller than KVQA in terms of the number of images and question-answer pairs, it is considerably bigger than the other knowledge-based VQA datasets.

Currently, \textbf{multimodal transformers} are the most successful systems for VQA and can be broadly classified into two types: single-stream and double-stream transformers. A good example of the former is VisualBERT \cite{li2019visualbert}, where the BERT architecture \cite{devlin2019bert} is used, adding visual features obtained by an object detector as input and using visio-linguistic pretraining tasks, such as image-text matching. OSCAR \cite{li2020oscar} also follows a very similar philosophy, adding object tags to the input and proposing different pretraining strategies. Among double-stream transformers, VilBERT \cite{lu2019vilbert} and LXMERT \cite{tan2019lxmert} use a dedicated transformer for each modality (text and image) to fuse them with a cross-modal transformer. Their differences lie mainly on some architectural choices and pretraining task selection \cite{bugliarello2020multimodal}. 

Regarding \textbf{OK-VQA systems}, multimodal transformers have also been used to provide implicit knowledge from pretraining tasks. For example, VilBERT uses a pretrained BERT to encode the questions, so it uses the implicit knowledge that BERT acquired during its pretraining. Additionally, VilBERT is further trained on Conceptual Captions \cite{sharma2018conceptual}, a large image-caption dataset from where additional knowledge can be acquired. Those multimodal transformers are the backbone of most models used for OK-VQA, which also use symbolic knowledge to bring some extra performance.

ConceptBert \cite{garderes2020conceptbert} was the first system to use multimodal transformers and symbolic knowledge for OK-VQA. It is based on a combination of a pretrained BERT to encode questions, a graph convolutional neural network to encode triples extracted from the ConceptNet knowledge graph \cite{speer2017conceptnet} and a multimodal transformer (VilBERT) to jointly represent and reason over image features and encoded question tokens. 

A similar approach was followed by KRISP \cite{marino2020krisp}, combining again a multimodal transformer with symbolic knowledge. In this case, the multimodal transformer, MM\textsubscript{BERT}, is based on VisualBert \cite{li2019visualbert} and initialized with the weights of a pretrained BERT. Additionally, authors built a knowledge graph fusing DBPedia \cite{auer2007dbpedia}, ConceptNet \cite{speer2017conceptnet}, VisualGenome \cite{krishna2017visual} and hasPart KB \cite{bhakthavatsalam2020dogs}. They used different image feature encoders and the question tokens to obtain a subset of the full graph relevant to the target question and image. Finally, using a graph convolutional neural network, they combined the symbolic and implicit knowledge to predict the final answer. 

Some recent approaches, named MAVEx \cite{wu2021multi} and RVL \cite{shevchenko2021reasoning} showed different ways to combine implicit and symbolic knowledge. MAVEx used a pretrained VilBERT to generate various candidate answers which were later validated using answer-specific knowledge retrieval. Authors used both textual and visual knowledge resources, including images searched using Google, sentences from Wikipedia articles, and concepts from ConceptNet. On the other hand, RVL trained the two-stream multimodal transformer LXMERT \cite{tan2019lxmert} with an auxiliary objective that aligned its representations with knowledge graph embeddings retrieved from ConceptNet and Wikidata. 



Integrating annotated captions, or other types of text related to the image, in multimodal systems benefit several multimodal challenges. Examples range from fake news detection \cite{kumari2021fakenews} to image classification tasks such as flower \cite{bae2020flower} and crisis \cite{ahmad2021crisis} classification. However, regarding the use of automatically \textbf{generated captions for VQA}, to the best of our knowledge, Mucko \cite{zhu2020mucko} and PICa \cite{yang2021pica} are the only systems that explore this idea. 
Mucko uses dense captions \cite{johnson2016densecap} to query a knowledge graph to extract relevant information to answer the question. The reported results on OK-VQA are well below the state-of-the-art. Dense captions describe different regions of an image using short sentences. Our method differs in the use of a single caption which is the input to the LM, and does not require neither knowledge graphs nor the use of OCR systems that have recently been integrated in some other VQA models \cite{singh2019towards,sharma2022phoc}.

On the other hand, PICa takes advantage of the implicit knowledge found in GPT-3 \cite{brown2020gpt3} via prompt-engineering. Instead of supervised fine-tuning, PICa adapts to the task with a few in-context examples during inference time using both captions and object tags to describe the image, defining the current state-of-the-art with an ensemble of GPT-3's and clever selection of those examples. However, GPT-3 is only accessible via OpenAI's paid API and it has limited functionalities.

\section{Implemented models}

\label{sec:models}
In this section we describe the implemented models. We use Pytorch \cite{paszke2019pytorch}, Pytorch Lightning and the Transformers library \cite{wolf2020transformers} for all the implementation work.

\subsection{Caption-based model (CBM)}
\label{sec:unibert}
Our caption-based model, denoted by CBM, is divided in two steps: (i) a caption generation system that generates a short description of a given image and (ii) a language model that takes this caption and a question in order to answer it.

We use \textbf{OSCAR} \cite{li2020oscar} to generate captions from images, a transformer encoder that produces state-of-the-art results on several multimodal tasks including image captioning. As it is common in multimodal transformers, OSCAR uses a pretrained object detector called FasterRCNN \cite{ren2015frcnn} to obtain region features from images and their respective labels. Both features and labels alongside manually annotated captions are then fed to the transformer during pretraining, following the work of \cite{anderson2018bottom}. The performance on image-captioning of both base and large models is similar, so we use OSCAR-base as our image-captioning system for all of our experiments. 

During OSCAR's fine-tuning step on image captioning, some of OK-VQA's test split images and gold captions are used. In order to ensure fairness and avoid any contamination in our experiments, we fine-tune a pretrained OSCAR model on image-captioning removing these instances from its training process.

For the second step, we explore two different language models: BERT, to perform comparative experiments with current multimodal transformers, and the T5 family, to explore the performance of LMs of increasing size.

\subsubsection{CBM\textsubscript{BERT}}

In this first approach, we use a pretrained \textbf{BERT-base} transformer encoder \cite{devlin2019bert} as our language model. We feed sequences of tokenized captions and questions $T^{(0)} = \{\mathbf{t}^{(0)}_{i} | i=1,\ldots,n_t\}$ to BERT, and take the output of the $[CLS]$ or first token of the sequence $\mathbf{t}^{(n_l)}_{1}$, where $n_t$ is the number of tokens in the sequence and $n_l$ is the number of transformer layers.

In order to fine-tune the language model for VQA tasks, we add a \textbf{classification head} to the $[CLS]$ embedding. Although VQA \cite{antol2015vqa,goyal2017vqav2} and OK-VQA \cite{marino2019ok} were defined with open-ended answers, recent models \cite{zhang2021vinvl,marino2020krisp} cast these tasks as classification problems, building a fixed vocabulary of answers from the training dataset. Following this trend, our classification head is a multilayer perceptron (MLP) with one hidden layer after $\mathbf{t}^{(n_l)}_1$. We define our MLP in Eq. \ref{eq:mlp}.

\begin{equation}
    \begin{split}
    \mathbf{h} &= \mathrm{LayerNorm}(\mathrm{GELU}(\mathbf{W}_h \mathbf{t}^{(n_l)}_1 + \mathbf{b}_h)) \\
    \mathbf{\hat{y}} &= \mathrm{Softmax}(\mathbf{W}_{\hat{y}} \mathbf{h} + \mathbf{b}_{\hat{y}})
    \end{split}
    \label{eq:mlp}
\end{equation}

We use a GELU activation function as well as layer normalization \cite{ba2016layernorm}. The trainable parameters are $\mathbf{W}_h \in \mathbb{R}^{d_{h} \times d_{h}}$, $\mathbf{b}_h \in \mathbb{R}^{d_{h}}$, $\mathbf{W}_{\hat{y}} \in \mathbb{R}^{d_{h} \times n_{label}}$ and $\mathbf{b}_{\hat{y}} \in \mathbb{R}^{n_{label}}$, where $n_{label}$ equals to the number of labels on a given classification task and $d_{h}$ equals to 768.

\subsubsection{CBM\textsubscript{T5}}

In our second approach, we use pretrained \textbf{T5} encoder-decoder transformers \cite{raffel2020t5}, as they are the state-of-the-art models for text-only question-answering tasks and are available in different sizes, ranging from 60M parameters to 11B. Following CBM\textsubscript{BERT}, we also feed sequences of tokenized captions and questions $T^{(0)} = \{\mathbf{t}^{(0)}_{i} | i=1,\ldots,n_t\}$ to the T5 model. Nevertheless, in this case we add text prefixes before each sentence, such as \textit{'caption:'} and \textit{'question:'}. This is mainly done to mimic the input prompts used during the pretraining process of the T5 model, helping the language model to better leverage what it has learnt before. Differently from BERT, T5 is a generative LM, so instead of classifying an answer, T5 produces it in an open-ended text generation manner. Thus we do not use any classifier head for this approach.

\subsection{Multimodal transformer (MM\textsubscript{BERT})}
\label{sec:multibert}
We compare our CBM\textsubscript{BERT} model with the multimodal transformer-based MM\textsubscript{BERT} \cite{marino2020krisp}, a variant of BERT that uses the question text and image region features as input. While BERT is designed to only process textual inputs, MM\textsubscript{BERT} adapts its embedding layer in order to be able to process features from images.

We use a FasterRCNN with a ResNeXt-152 \cite{xie2016resnext} as its backbone to extract a total of $n_v$ region features $\mathbf{V} = \{\mathbf{v}_{1}, \ldots, \mathbf{v}_{n_v}\}$ per image. Each of these $\mathbf{v}_i \in \mathbb{R}^{d_v}$ features represents an object that appears in the image, where $d_v$ equals to 2048. $\mathbf{V}$ lacks the positional information between objects, which can be solved concatenating the corresponding bounding box coordinates to each feature. Upon some initial experiments, we concluded that this extra information does not improve performance in any of VQA 2.0 and OK-VQA. 
We use MMF Multimodal Framework \cite{singh2020mmf} to extract the image region features that are fed into MM\textsubscript{BERT}. 

In order to allow for easier comparison between our CBM and MM\textsubscript{BERT} we use the output representation for $[CLS]$ to feed into the classification multilayer perceptron (see Section \ref{sec:unibert}). Note that this is slightly different from the original MM\textsubscript{BERT} \cite{marino2020krisp}, which uses the average of all token representations in the last transformer layer.

\subsection{Question-only baseline (Q\textsubscript{BERT})}
In order to assess the contribution of captions, we also trained a model which only had the question in the input, without any information about the image or caption, denoted as Q\textsubscript{BERT}. This model can be seen as an ablation of CBM\textsubscript{BERT}. 

\subsection{Loss function}
Contrary to previous works in VQA, we do not use binary cross-entropy loss for our classification models, as initial experiments showed that cross-entropy loss with soft labels (SCE) converges faster with similar results. SCE loss is defined in Eq. \ref{eq:sce}, where $\mathbf{y}$ is the ground truth vector with probabilities proportional to the VQA evaluation metric (Eq. \ref{eq:vqa_accuracy}) assigned to each class.

\begin{equation}
 \mathcal{L}_{SCE}(\mathbf{y}, \mathbf{\hat{y}}) = - \mathbf{y} \cdot \log \mathbf{\hat{y}}
 \label{eq:sce}
\end{equation}

Regarding CBM\textsubscript{T5}, we fine-tune this generative model via teacher forcing using cross-entropy loss. Therefore, the model learns to map each input sequence with its respective target sequence. However, training the model using the teacher forcing paradigm causes a discrepancy with the human annotations, as each question in OK-VQA has multiple valid answers. We fix this by randomly choosing a target sequence on each epoch. Initial experiments also showed that randomly choosing an answer from all annotated answers is slightly detrimental, as some answers are not spelled correctly, are empty strings or do not make sense as an answer. Therefore, during training we exclude answers that do not obtain a full score in the VQA score, that is, we choose answers that are annotated by at least two annotators on a given question\footnote{If a question does not have answers that fulfill this rule, the question is discarded from training, which amounts to a total of 112 instances in the OK-VQA's training split.}.

\begin{figure*}[t]
\includegraphics[width=\linewidth]{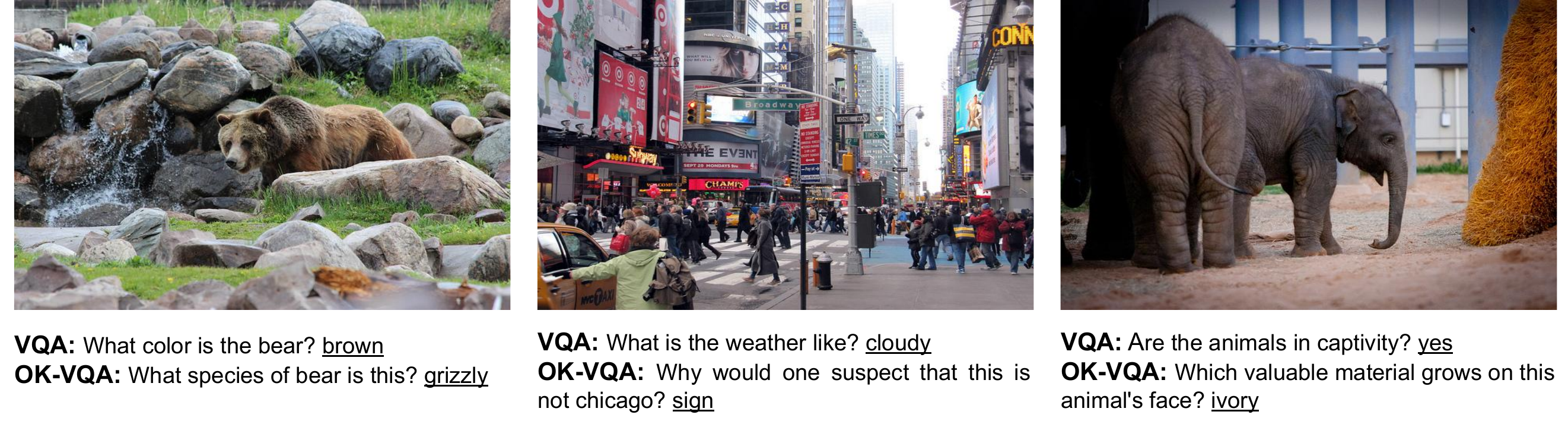}
\caption{Some examples of VQA 2.0 and OK-VQA datasets for the same images. VQA questions are about image contents, while OK-VQA questions require outside knowledge. 
} \label{fig:dataset}
\end{figure*}

\section{Datasets}
\label{sec:datasets}
The main dataset for our experiments is OK-VQA \cite{marino2019ok}, since it allows us evaluating the usage of the implicit knowledge of LMs in a multimodal task. But we also run experiments on the VQA 2.0 dataset \cite{goyal2017vqav2} with a double motivation: (i) to use it as additional pretraining before applying the model to OK-VQA; (ii) to analyze the performance differences among models on a knowledge-based VQA dataset and a standard VQA dataset. 
Examples of both datasets can be found in Figure \ref{fig:dataset}.

\subsection{VQA 2.0}
This dataset contains open-ended questions about images 
where questions focus mainly on identifying objects in the image and their attributes, detecting relations between them, as well as counting those objects. The dataset is composed of 204K images taken from the COCO dataset \cite{lin2014coco} and 1.1M questions, each question having 10 (possibly repeated) annotations as accepted answers. Following the classification setting of VQA tasks, which is currently the dominant paradigm, VQA 2.0 has 3129 different possible answers, extracted from the most frequent answers of the training split. 

VQA 2.0 is divided in three splits named train, dev and test. Some of the images from the development split of VQA 2.0 are reused in OK-VQA's test split. So, in order to avoid any contamination, we do not use the VQA 2.0 dev set for any training or hyper-parameter tuning.

Antol et al, \cite{antol2015vqa} proposed a standard evaluation metric for VQA tasks where a system answer is considered totally correct if it appears at least three times in the ten ground-truth annotations. Considering that a given answer appears $x$ times in a question's annotations, this accuracy metric is defined in Eq. \ref{eq:vqa_accuracy}.

\begin{equation}
    \mathrm{acc} = \min \left( \frac{x}{3}, 1 \right)
    \label{eq:vqa_accuracy}
\end{equation}

\subsection{OK-VQA}
The OK-VQA dataset is built upon 14,031 images from the COCO dataset and 14,055 crowd-sourced questions. Each question has ten annotated answers (possibly repeated), and the evaluation metric is the same as in  VQA 2.0  (Eq. \ref{eq:vqa_accuracy}).
As a knowledge-based VQA dataset, OK-VQA requires outside knowledge to answer the questions. However, this outside knowledge is neither provided nor identified, i.e. there is not a list of available knowledge sources for this task, making the task more challenging.

There are two versions of this dataset, depending on how the stemming of the answers provided by the crowd-sourcers is handled. The stemming used in OK-VQA v1.0 results in some ``non-word'' answers (such as ``poni tail'' instead of ``pony tail''). OK-VQA v1.1 applied a different stemming algorithm, resulting in a more coherent answer vocabulary. We use OK-VQA v1.1 through our experiments.

\section{Experiments and results}
\label{sec:experiments}

This section provides results of the models defined in Section \ref{sec:models} and compare them with the state-of-the-art. 

\subsection{Experimental settings}

We use the same hyperparameters as \cite{marino2020krisp} for fine-tuning CBM\textsubscript{BERT}, MM\textsubscript{BERT} and Q\textsubscript{BERT} models both in VQA 2.0 and OK-VQA tasks. We train our models for 88K steps using AdamW optimizer \cite{loshchilov2018adamw}. Our batch size is of 56 with a maximum learning rate of $5\cdot10^{-5}$ following a cosine schedule with a linear warmup of 2K steps.

Regarding CBM\textsubscript{T5}, there are 5 different T5 models that vary on size. They range from 60M to 11B parameters and we show the performance of all five models on OK-VQA. To do so, we have chosen to keep the same hyperparameters as before with the following changes:

\begin{itemize}
    \item As models of different sizes need different amounts of training steps in order to converge, we propose the following methodology. We use $20\%$ of the training instances to define a validation split, train the models using the remaining $80\%$ for 20K steps and decide the final number of steps by taking the step with the best VQA score in the validation split. This process is done three times using the same validation split. After that, we compute the average number of steps of all three runs as our final number of training steps.
    \item As the number of training steps varies among different model sizes, we have decided to use a fixed learning rate of $5\cdot10^{-5}$ during the fine-tuning process, without any learning-rate scheduler that depends on warmup steps or total number of training steps.
\end{itemize}

All experiments regarding classification models have been run in a single GPU with 12GB of vRAM and their runtimes are at most of 12 hours. Regarding the much larger CBM\textsubscript{T5} we used up to 4 NVIDIA A100 GPUs (each with 80 GB of vRAM), changing both hardware and hyper-parameters to keep the same effective batch size across model sizes, and used DeepSpeed's ZeRO Stage 2 optimization algorithm with CPU offload \cite{rajbhandari2020zero} when fine-tuning the biggest model. However, their runtimes are at most of 4 hours, as less training steps are needed for CBM\textsubscript{T5}, compared to the rest of our models.


In order to get consistent results we make each experiment three times and provide the mean VQA score and standard deviation in all of our results.

\subsection{Results for images vs. captions}

\begin{table}[t]
\centering
\begin{tabular}{lccc}
\toprule
$\mathrm{Model}$ & Score & + VQA pretraining & Parameters \\ \midrule
Q\textsubscript{BERT}            &     $21.2$  \small$\pm0.2$ &     $23.0$  \small$\pm0.2$ & 112M  \\
MM\textsubscript{BERT}           &     $29.6$  \small$\pm0.6$ &     $35.7$  \small$\pm0.3$ & 114M  \\
CBM\textsubscript{BERT}  (ours)  & $\bf{32.5}$ \small$\pm0.4$ & $\bf{36.0}$ \small$\pm0.4$ & 112M \\ \bottomrule
\end{tabular}
\caption{Performance on OK-VQA for three classification models (respectively, question only, image-based and caption-based) without and with additional pretraining on VQA 2.0. Mean VQA score and standard deviation across 3 different runs.}
\label{tab:modalities}
\end{table}

Table \ref{tab:modalities} shows the results for the three models presented in Section \ref{sec:models}, which share the same architecture, size and initial parameters. We show the results for the models fine-tuned on OK-VQA, as well as the same models which have been previously fine-tuned on VQA 2.0. 

We observe that the sole use of questions (Q\textsubscript{BERT}) offers poor performance compared to the other two systems, achieving up to 13 points less accuracy. This shows that having any representation of the image (captions or image region features) is key to answer questions correctly. This is further justified comparing the improvement that VQA pretraining entails, as Q\textsubscript{BERT} improves less than 2 points, whereas the other two improve their accuracy between 4-6 points.

\paragraph{Contribution of captions}
When we compare the performance of CBM\textsubscript{BERT} and MM\textsubscript{BERT}, we see that, when there is no visio-linguistic pretraining involved, CBM\textsubscript{BERT} performs better in OK-VQA. However, when we pretrain these models in a similar multimodal task like VQA 2.0, their accuracy increases by 4-6 points and both obtain similar performance. As OK-VQA's training is comparatively smaller (9K instances vs. VQA's 410K instances), we hypothesize that training MM\textsubscript{BERT} on OK-VQA is not enough to adapt the model to the new input modality. However, as CBM\textsubscript{BERT} uses only text, the fine-tuning with such small training is more effective.

\subsection{T5 and larger models}
As T5 ha been pre-trained on several question answering tasks, we directly fine-tune it on OK-VQA alone.

\begin{table}[t]
	\begin{minipage}{0.48\linewidth}
		\centering
		\begin{tabular}{lcc}
        \toprule
        Model & Score & Parameters  \\ \midrule
        CBM\textsubscript{T5-Small}      &      $29.2$ \small$\pm0.2$ & 60M \\ 
        CBM\textsubscript{T5-Base}       &      $36.1$ \small$\pm0.5$ & 220M \\ 
        CBM\textsubscript{T5-Large}      &      $40.8$ \small$\pm0.4$ & 770M \\ 
        CBM\textsubscript{T5-3B}         &      $44.0$ \small$\pm0.7$ & 3B \\ 
        CBM\textsubscript{T5-11B}        & $\bf{47.9}$ \small$\pm0.2$ & 11B  \\  \bottomrule
        \end{tabular}
        \caption{Performance on OK-VQA of our generative CBM\textsubscript{T5} models.}
        \label{tab:t5results}
	\end{minipage}\hfill
	\begin{minipage}{0.48\linewidth}
		\centering
		\includegraphics[width=\linewidth]{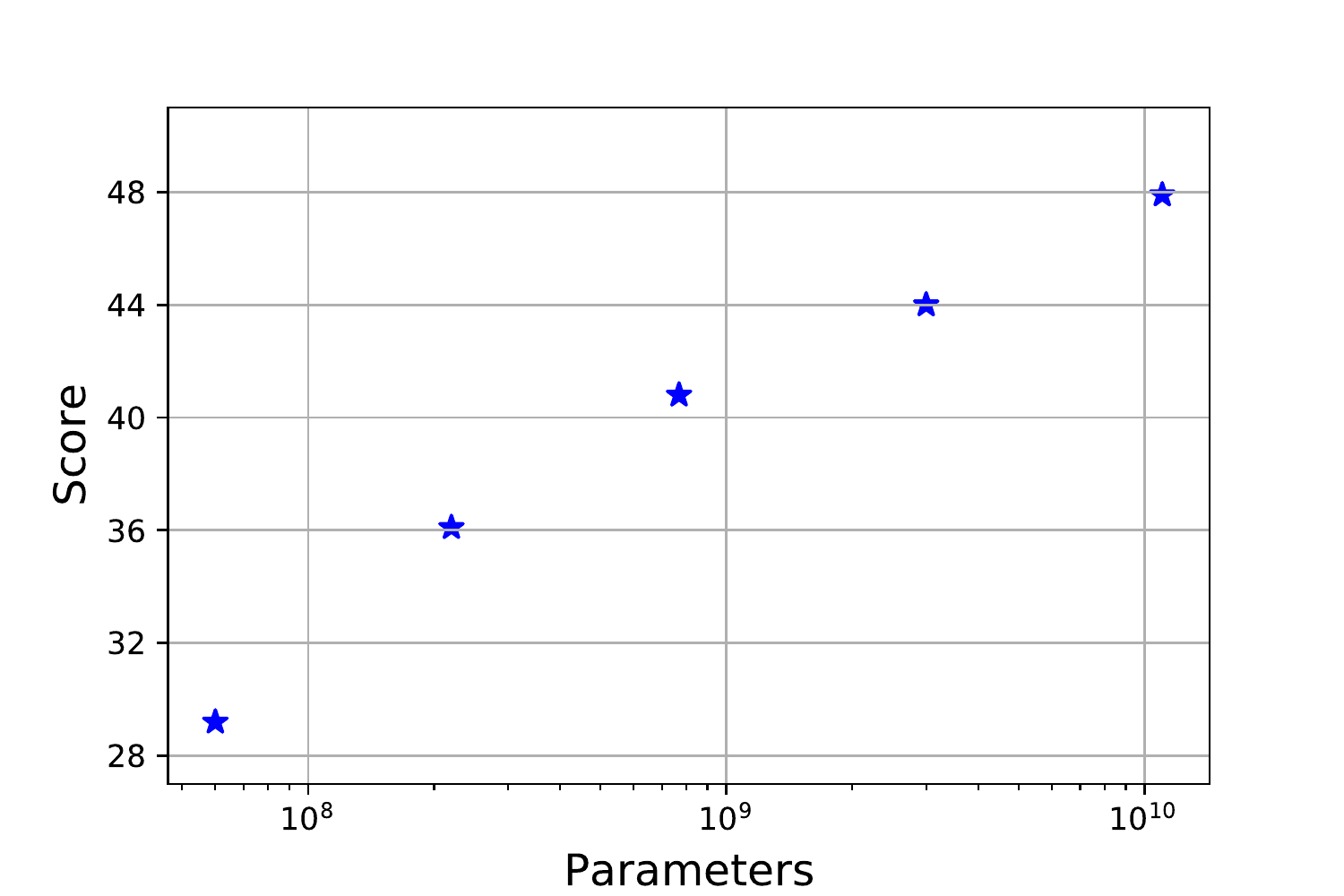}
		\captionof{figure}{Correlation between the size of CBM\textsubscript{T5} models and their performance. The horizontal axis is in logarithmic scale.}
		\label{fig:t5trend}
	\end{minipage}
\end{table}

In Table \ref{tab:t5results} we show the results of five differently sized CBM\textsubscript{T5} models on OK-VQA. Note that our T5-Base model obtains results comparable to our BERT-base model pre-trained on VQA 2.0. This was expected, as both models have been pre-trained with VQA datasets and both have comparable model sizes, T5-base being composed of two BERT-base encoder and decoder. 

The results in Table \ref{tab:t5results} are plotted in Figure \ref{fig:t5trend}, showing that the size of our models is logarithmically proportional to its score, which follows the scaling laws mentioned in \cite{kaplan2020scaling}. This trend is followed even by our biggest model and does not seem to slow down yet.
These results show the importance of the model's capacity in the results. All models have been pretrained with the same corpus and downstream tasks, but the difference in size helps bigger models to better leverage the information learnt from that corpus in order to incorporate the external knowledge needed to solve OK-VQA.  Our largest model performs much better than the multimodal model.

In fact, it is not clear whether larger multimodal models could match our largest text-only caption based model. We cannot currently test this hypothesis, as, to the best of our knowledge, there are no publicly available multimodal transformers with comparable numbers or parameters. Still, we hypothesize that in the case of knowledge-intensive datasets such as OK-VQA, current multimodal transformers \cite{lu2019vilbert, li2019visualbert, tan2019lxmert} will underperform our system, as the only textual data fed to these models during their pretraining is mostly composed by captions or small descriptions attached to images. This means that these models only see a limited vocabulary from a limited corpus, compared to the rich, diverse and much larger corpora used to build models such as T5.


\color{black}
\subsection{Comparison with the state of the art}



In Table \ref{tab:sota}, we show the results of various state-of-the-art models in three groups: 1) classification models based on multimodal transformers, which additionally include the usage of symbolic knowledge; 2) GPT-3 based generative models that use in-context learning; 3) our caption-based models. 

\begin{table}[t]
\centering
\begin{tabular}{lccc}
\toprule
Model & Score &  & Parameters \\ \midrule
ConceptBERT \cite{garderes2020conceptbert} * & 31.4 & ({\small +sym.} 33.7) & 348M \\
MAVEx \cite{wu2021multi} & 35.2 & ({\small +sym.} 41.4) & 353M \\
KRISP \cite{marino2020krisp} & 37.1 & ({\small +sym.} 38.9) & 116M \\ 
RVL \cite{shevchenko2021reasoning} *\dag & 37.3 & ({\small +sym.} 39.0) & 208M \\ \midrule
PICa-Base \cite{yang2021pica} & 42.0 & ({\small +tags} \ 43.3) & 175B \\
PICa-Full \cite{yang2021pica} \small(Ensemble) & 46.9 & ({\small +tags} \ \textbf{48.0}) & 175B \\ \midrule
CBM\textsubscript{BERT} (ours) & 36.0 & & 112M \\ 
CBM\textsubscript{T5-11B} (ours) & \textbf{47.9} & & 11B \\ \bottomrule
\end{tabular}
\caption{Comparison to the state-of-the-art on OK-VQA. +sym. stands for systems additionally using symbolic knowledge, and +tags for the additional use of object tags. Results of models marked with * are in OK-VQA v1.0 and \dag \ specifies contaminated results (see main text).}
\label{tab:sota}
\end{table}

The state-of-the-art classification models like KRISP \cite{marino2020krisp}, MAVEx \cite{wu2021multi} and RVL \cite{shevchenko2021reasoning} show similar results on the implicit-only versions of their models, even though they are based on different multimodal transformers and pretraining tasks. Note that RVL has a contamination issue as images from OK-VQA's test split were used to train their multimodal transformer. We also observe that using symbolic knowledge improves the results around 2 points, the exception being MAVEx   which combines knowledge found in ConceptNet \cite{speer2017conceptnet}, Wikipedia and Google Images \footnote{This result is obtained with an ensemble of 3 MAVEx models that share the same multimodal transformer. A unique MAVEx model achieves an accuracy of 40.3\%.}. 

PICa \cite{yang2021pica} takes advantage of GPT-3 \cite{brown2020gpt3} to define a new state-of-the-art in a generative manner using in-context learning. Its base results (PICa-Base) already surpass the ones seen before without any need of symbolic knowledge. An ensemble of 5 GPT-3 models and a clever selection of annotated examples from the training data to build the input prompt further improves its results (PICa-Full). Table \ref{tab:sota} reports two results for each PICa model: the results using automatically generated captions alone (like us), and the results when also using object tags automatically obtained from the image, which slightly improve the results.

Our CBM\textsubscript{BERT} system performs on par with the multimodal transformers, which is remarkable, since we do not use directly any visual features in our models and only use the caption. Note that all those systems have models of comparable size. When scaling up our generative models, we see that CBM\textsubscript{T5-11B} outperforms current multimodal models by a large margin and is on par with the results obtained by PICa-Full. Indeed, CBM\textsubscript{T5-11B} achieves slightly better results than the PICa version which uses captions alone, even if our model is 15 times smaller.


\section{Analysis}

In this section we perform additional experiments. We first contrast the results on OK-VQA with those obtained in VQA 2.0, discussing the reasons for the different performance. We then combine our text only model with its counterpart multimodal model to analyze if they are complementary. Afterwards, we compare the performance of CBM\textsubscript{BERT} with manually annotated captions or the ones generated by OSCAR \cite{li2020oscar}. Finally, we present some qualitative analysis.

\subsection{Results on VQA 2.0}

\begin{table}[t]
\centering
\begin{tabular}{lc}
\toprule
Model & Score  \\ \hline
MM\textsubscript{BERT}      & \textbf{65.8} \\
\midrule
PICa-Full & 56.1 \\
CBM\textsubscript{BERT}  (ours) & 59.6 \\
\bottomrule
\end{tabular}
\caption{Performance on the dev split of VQA 2.0 of the multimodal model MM\textsubscript{BERT} and two text-only models: PICa-Full and CBM\textsubscript{BERT}.}
\label{tab:vqa}
\end{table}

Even though both unimodal and multimodal methods perform similarly in OK-VQA, we observed a different trend in VQA 2.0. Table \ref{tab:vqa} shows that CBM\textsubscript{BERT} obtains $59.6$, while MM\textsubscript{BERT} achieves 6 points more. We think this is due to the information loss when converting an image into a caption, as relevant information that is needed to answer the question can be lost. This is specially important for VQA 2.0, where the questions refer directly to image contents, spatial relations and object attributes (see Figure \ref{fig:dataset}). A similar behavior can be observed for PICa \cite{yang2021pica}. Interestingly, PICa also uses object tags to minimize the information loss when verbalizing the image, but it does not perform as well as our system. Even with 1000 times less parameters, our CBM\textsubscript{BERT} outperforms PICa, showing the importance of fine-tuning in contrast to in-context-learning, specially when large training data is available, as in VQA 2.0.

The difference between VQA and OK-VQA performances suggests that captions contain enough information to effectively use the implicit knowledge of language models for knowledge-intensive multimodal tasks like OK-VQA, but that in datasets where the answer can be found in the image, multimodal models preferable. 

\subsection{Combining visual information and captions}
\label{sec:combining}
Given the different nature of the inputs, we wanted to check whether CBM\textsubscript{BERT} and MM\textsubscript{BERT} are complementary. Our hypothesis is that the former can take advantage of the implicit knowledge acquired by the language model, whereas the latter has access to more fine-grained information found in image regions. Therefore, we define two different approaches to check how they complement each other.

\textbf{Early fusion.} For each question we feed both caption and image features alongside the question to the language model. This system can be seen as a MM\textsubscript{BERT} which processes a multimodal input composed by a question (text), a caption (text) and image region features. We initialize the weights of this model with the weights of the base language model (BERT-base) and fine-tune it on the target train data.

\textbf{Late fusion.} We train CBM\textsubscript{BERT} (Section \ref{sec:unibert}) and MM\textsubscript{BERT} (Section \ref{sec:multibert}) separately, each of them with their corresponding inputs, and combine their outputs in inference time to obtain the final answer. The combination is done by multiplying output probabilities of both models for each class and taking the answer with the highest value. We show their performance in Table \ref{tab:fusion}.

These fusion models improve the performance of both CBM\textsubscript{BERT} and MM\textsubscript{BERT} by 2-3 points in almost all cases. The only case where there is no improvement comparing to CBM\textsubscript{BERT} is in the early fusion without VQA pretraining. This may be caused again by the small training split of OK-VQA, causing difficulties to learn how to ground textual and visual modalities. However, this is solved when VQA pretraining is added to the model, increasing vastly the amount of data seen by the models and showing similar performance on both early and late fusion models. 

Additionally, we also observed the complementarity of both modalities in the VQA dataset. Early fusion obtains $67.8\%$ and late fusion $67.7\%$ in the dev split of VQA 2.0, improving the performance of MM\textsubscript{BERT} by 2 points. The results validate our hypothesis, showing that image region features and captions are complementary in this setting.

\begin{table}[t]
\centering
\begin{tabular}{lcc}
\toprule
$\mathrm{Model}$ & Score & + VQA pretraining \\ \midrule
Early Fusion    &     $32.5$  \small$\pm0.4$ &     $38.2$  \small$\pm0.8$ \\
Late Fusion     &     $34.0$  \small$\pm0.4$ &     $\bf{38.6}$  \small$\pm0.2$ \\ \bottomrule
\end{tabular}
\caption{Performance on OK-VQA for early and late fusion models.} 
\label{tab:fusion}
\end{table}

\subsection{Ground truth captions}

In order to measure the effects of the image captioning system to our proposed CBM model, we check the gap of performance between the use of generated captions and gold captions. As OK-VQA is built upon images from the COCO dataset \cite{lin2014coco}, each image has five different annotated captions. We use these captions and fine-tune CBM\textsubscript{BERT} on OK-VQA without VQA pretraining following the same experimental settings. 
We repeat this experiment three times, as it is done through the entire work. On each run we select a different set of captions, that is, for each image we just choose one gold caption randomly and use it during the entire training process. As we also have several captions in all of OK-VQA's test split, we test each fine-tuned model three times following the same caption selection process. 

Table \ref{tab:modalities} already shows that we achieve an accuracy and standard deviation of $32.5 \pm 0.4$ using generated captions on OK-VQA's test split. However, when we use gold captions we get an average accuracy of $32.3 \pm 0.3$ in all of our runs. In both cases we obtain similar results, showing that captions generated by OSCAR contain enough information for CBM\textsubscript{BERT} to perform comparably on this specific task.


\subsection{Qualitative Analysis on OK-VQA}

\begin{figure*}[t]
\includegraphics[width=\linewidth]{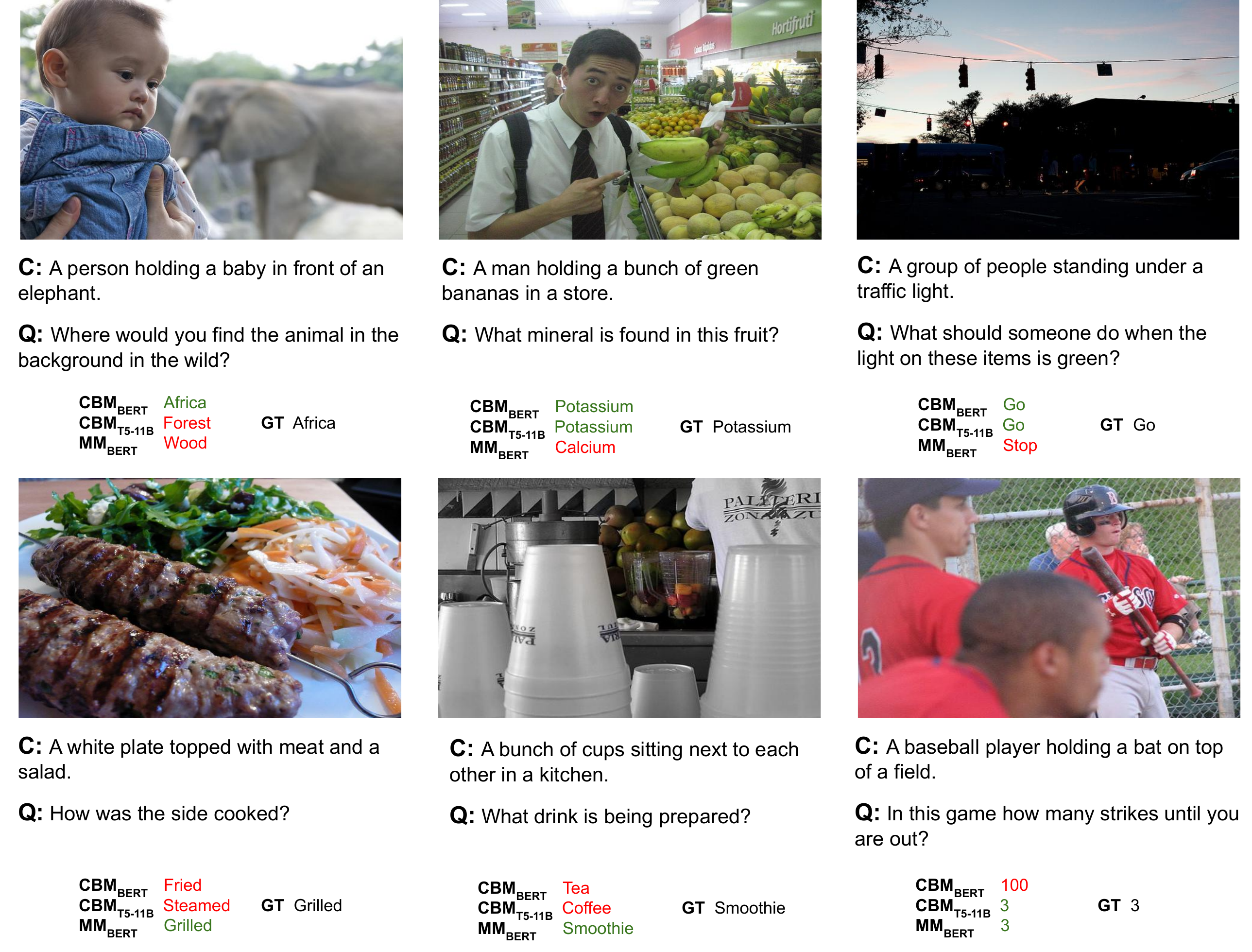}
\caption{Examples of OK-VQA questions where only one between CBM\textsubscript{BERT and MM\textsubscript{BERT}} answers correctly according to the ground truth (GT). We also show answers given by CBM\textsubscript{T5-11B} for further comparisons. C refers to captions generated by OSCAR. Correct answer in green, incorrect in red.} \label{fig:qualitative}
\end{figure*}

Both CBM\textsubscript{BERT} and its multimodal counterpart perform similarly (see Table \ref{tab:modalities}), but in 38.7\% of the test examples their output differs and only one of them is correct. Figure \ref{fig:qualitative} shows some OK-VQA test examples together where the outputs of CBM\textsubscript{BERT} and MM\textsubscript{BERT} with VQA pretraining differ. We also add answers from CBM\textsubscript{T5-11B} for further comparisons.


Starting with the top-left example, CBM\textsubscript{BERT} can infer that elephants are native to Africa whereas MM\textsubscript{BERT} does not. In fact, the generated caption includes the information that the animal found in the image is an elephant, performing the first step needed to answer the question. This way, the LM can focus on using its implicit knowledge in order to answer correctly. CBM\textsubscript{T5} generates 'forest' as an answer. Although the answer may be considered as valid to us, the answer is not within the list of ground truth answers, making it incorrect.
The other two examples in the top row behave similarly. The caption facilitates the grounding between the question and the image. Whenever a question refers to the image (``this fruit'' and ``these items''), if the caption already mentions these objects (``bananas'' and ``traffic light'', respectively), the LM seems to better leverage its implicit knowledge and reasoning capabilities to answer the question. The top-right example is interesting in this regard. While the image shows red traffic lights, the question asks about the effects of green lights. This may trick MM\textsubscript{BERT} into answering the effect that red lights produce, not the green ones. 

The bottom row of Figure \ref{fig:qualitative} shows two examples where the caption does not give enough information to infer the answer for CBMs. In the first case CBMs cannot decide whether the meat is steamed, fried or grilled by only examining the caption, while MM\textsubscript{BERT} does have access to visual cues of the image, where we can see that the meat is grilled. This also happens in the second example, as the caption does not specify any ingredient of the beverage while we can see fruits in the image. The rightmost example illustrates an example where the caption does support the inference, but where our BERT based CBM gets it wrong. With the given caption, ``this game'' refers to baseball, however, CBM\textsubscript{BERT} is unable to infer that three strikes are enough for a strikeout whereas both CBM\textsubscript{T5-11B} and MM\textsubscript{BERT} manage to give the correct answer.

All in all, these examples support our hypothesis that visual features and captions are complementary. They also show that our system has some advantages regarding the interpretability of the system, specially in the cases our method is wrong. In some cases like the two leftmost examples in the bottom row, the object or feature needed to answer the question is missing from the caption. In other cases, the required information is in the caption, but the inference is erroneous. 

\section{Conclusions}

In this paper we present a VQA system which describes images with a caption to then work only with textual data. We show that such a system performs surprisingly well in OK-VQA, where the questions cannot be answered with the image alone, requiring access to external knowledge. Our analysis indicates that the loss of information when summarizing the image into a caption is compensated by the better inference ability of text-only pretrained language models. We also show the importance of a language model's capacity when leveraging the implicit knowledge found in it, achieving state-of-the-art results, outperforming current multimodal models by a large margin and matching a 15-times larger ensemble model. Compared to multimodal models, orders of magnitude bigger text-only LMs are available, which we show to be an advantage for knowledge-intensive tasks. In the future we would like to explore whether richer descriptions of images might improve results further, and whether large text-only language models can still benefit from incorporating symbolic knowledge graphs. 

\bibliography{custom} 
\bibliographystyle{elsarticle-num}

\end{document}